\documentclass[conference]{IEEEtran}
\IEEEoverridecommandlockouts
\usepackage{cite}
\usepackage{amsmath,amssymb,amsfonts}
\usepackage{algorithmic}
\usepackage{algorithm}
\usepackage{graphicx}
\usepackage{textcomp}
\usepackage{xcolor}
\usepackage{tikz}
\usetikzlibrary{arrows.meta,positioning,calc,fit}
\def\BibTeX{{\rm B\kern-.05em{\sc i\kern-.025em b}\kern-.08em
    T\kern-.1667em\lower.7ex\hbox{E}\kern-.125emX}}
\begin{document}

\newcommand{\eref}[1]{(\ref{#1})} 
\newcommand{\sref}[1]{Sec. \ref{#1}} 
\newcommand{\figref}[1]{Fig.\ref{#1}} 
\newcommand{\tref}[1]{Table \ref{#1}} 
\newcommand{\aref}[1]{Algorithm \ref{#1}} 

\newboolean{include-notes}
\setboolean{include-notes}{true}
\newcommand{\rhnote}[1]{\ifthenelse{\boolean{include-notes}}%
 {\textcolor{blue}{RH: #1}}{}}

\title{Adaptive Surface Cleaning under Uncertain Cleaning Response
}

\author{Ruichen Li, Rachel Holladay
\thanks{All authors are with the General Robotics, Automation, Sensing,
and Perception (GRASP) Laboratory, University of Pennsylvania, USA $\langle$\texttt{spencrli, rhollada}$\rangle$\texttt{@seas.upenn.edu}.}}

\maketitle



\section{Introduction}
\label{sec:intro}

Surface cleaning is a contact-rich manipulation task that requires a robot to decide not only where contamination exists, but also how to interact with it. Because different contaminants demand varying motions, applied forces, or repeated passes, effective cleaning requires determining both the spatial organization of the cleaning plan and the parameters of the cleaning actions during execution.

Making these decisions adaptively is challenging because the contamination response to cleaning varies across the surface and cannot be known in advance. Two visually-similar regions may respond differently to the same wiping action due to unobservable factors like contaminant composition, adhesion, drying, or surface properties. Consequently, the robot must instead infer these reactions from physical interaction, using observed outcomes to continuously adapt its subsequent actions.

Prior work has either learned cleaning transitions from action outcomes or assumed analytical models to support planning and adaptation~\cite{hess2011learning,leidner2019cognition,elliott2018robotic,lew2022robotictablewipingreinforcement}. Other approaches focus on transferring demonstrated cleaning motions across tools and surfaces or adapting contact and force during execution~\cite{elliott2017learning,bilaloglu2025tactile,tsuji2025adaptive}. These methods address important components of the cleaning task, but do not explicitly account for uncertainty in the spatially-varying cleaning response during planning.

To reason about this uncertainty, we treat the unknown cleaning response as a latent component of the surface state. Belief-space planning provides a natural framework for decision making when relevant state variables are uncertain or unobservable~\cite{platt2010belief,qiu2020latent}, while physical interaction can provide observations that refine latent-state estimations and support adaptive replanning~\cite{hsiao2010task}. We therefore formulate adaptive cleaning as a partially observable Markov decision process (POMDP). Each surface region contains a visible contamination state and a latent response state. We instantiate the latent response using a scalar force-response surrogate and maintain a joint per-cell belief over the two states. After each execution, the discrepancy between predicted and observed contamination updates this belief, guiding subsequent cleaning motions, forces, and passes.

To solve this POMDP, we propose a belief-aware planning framework that alternates surface observation, belief updating, geometry-based action generation, plan evaluation, execution, and replanning (\figref{fig:approach_overview}). Candidate plans are constructed from alternative spatial clusterings of contamination and parameterized over discrete force levels and pass counts. Preliminary grid-world simulations validate the closed-loop planner under simplified cleaning dynamics. Future work will extend this evaluation to physical robot execution with visual perception.

\section{Problem Formulation}
\label{sec:problem_formulation}

We frame adaptive surface cleaning as a partially observable Markov Decision Process (POMDP)
\begin{equation}
    \mathcal{P}
    =
    \left(
        \mathcal{S},
        \mathcal{A},
        T,
        \mathcal{Z},
        O,
        C
    \right),
    \label{eq:pomdp}
\end{equation}
where $\mathcal{S}$ is the state space, $\mathcal{A}$ is the action space, $T$ is the transition model, $\mathcal{Z}$ is the observation space, $O$ is the observation model, and $C$ represents the cost associated with executing actions.

\subsection{State Space}
\label{sec:state_space}

We consider contamination distributed over a bounded planar surface that is discretized into a  grid:
\begin{equation}
    \mathcal{G}=\{1,\ldots,N\},
    \label{eq:surface_grid}
\end{equation}
where each cell $j\in\mathcal{G}$ indexes a spatial region of the surface (\figref{fig:approach_overview}).
We associate each cell $j$ with two state variables: a contamination state $x_j\in\mathbb{R}_{\geq 0}$ and a cleaning-response state $\theta_j\in\Theta$.
The contamination state $x_{j}$ may vary spatially, is observed only through sensor-derived measurements, and $x_{j}=0$ corresponds to a clean cell. The latent response state $\theta_{j}$ captures the spatial variation in how contamination responds to cleaning actions, which is not directly observable and must be inferred from the observed outcomes of actions. It is an abstract parameterization of the local cleaning dynamics which represents the unknown physical quantities that determine how the contamination state changes under a cleaning action, e.g., through a transition of the form 
\begin{equation}
    x_j'\sim p(x_j'\mid x_j,a;\theta_j).
\end{equation}

We intentionally do not prescribe the dimension or physical interpretation of $\Theta$, since different classes of contamination may require different transition models and response parameters. Thus, $\theta_j$ represents whatever latent properties are required to parameterize the underlying cleaning response. Currently, we adopt a scalar surrogate for this latent response to obtain a tractable belief and transition model (\sref{sec:belief_modeling}). 

The state of cell $j$ is $s_j=(x_j,\theta_j)$ and the complete surface state is $s=(x,\theta)\in\mathcal{S}$ where $x=\{x_j\}_{j\in\mathcal{G}}$ and $\theta=\{\theta_j\}_{j\in\mathcal{G}}$.

\subsection{Observation Model}
\label{sec:observation_model}

The sensor observation is an image $o$ of the surface that we partition into the grid $\mathcal{G}$.
Let $o_j$ denote the image crop corresponding to cell $j$. A perception function $g$ maps each image cell to a normalized appearance score
\begin{equation}
    z_j=g(o_j)\in[0,1],
    \label{eq:appearance_observation}
\end{equation}
with the zero-anchor property
\begin{equation}
    g(o_j)=0
    \quad\Longleftrightarrow\quad
    x_j=0.
    \label{eq:observation_zero_anchor}
\end{equation}
The complete processed observation is $z=\{z_j\}_{j\in\mathcal{G}}\in\mathcal{Z}$.

Currently, we assume that $z$ is observed without noise and do not model uncertainty introduced by sensing, image registration, or the perception function $g$. Accordingly, $z_{j}$ is treated as a deterministic observation of the visible contamination state $x_{j}$, while the latent response state $\theta_j$ remains unobserved.

\subsection{Action Space and Cost Function}
\label{sec:action_space}

We define the action space as a finite set of parameterized cleaning and repositioning controllers,
\begin{equation}
    \mathcal{A}
    =
    \mathcal{A}_{\mathrm{wipe}}
    \cup
    \mathcal{A}_{\mathrm{spiral}}
    \cup
    \mathcal{A}_{\mathrm{scratch}}
    \cup
    \mathcal{A}_{\mathrm{move}}.
    \label{eq:action_space}
\end{equation}
We associate each action $a\in\mathcal{A}$ with a cost $c(a)$ determined by its parameters, and evaluate candidate plans by summing the costs of their constituent actions. To penalize larger applied forces and longer motions in cost functions, we introduce a force-dependent cost factor and measure motion length using Euclidean distance in grid coordinates. Manually specified weights balance the contributions of path length, covered area, and travel distance. The exact action parameterizations and action-specific cost functions are provided in Appendix \ref{app:action_definitions}. Actions will be implemented on the real robot using joint position and Cartesian impedance controllers.

\subsection{Transition Model}
\label{sec:transition_model}

We assume that cleaning induces a stochastic state transition whose effect on the contamination state depends on the latent cleaning-response state $\theta$. Given the joint surface state $s$, the transition model is
\begin{equation}
T(s'\mid s,a)
=
p(s'\mid s,a).
\label{eq:state_transition}
\end{equation}
The transition dynamics are initially unknown to the planner because the cleaning response component $\theta$ is not observable. Interaction with the surface therefore provides information not only about the resulting contamination state but also about how the surface responds to cleaning actions.

We assume spatially independent cell transitions conditioned on the applied action:
\begin{equation}
    p(s'\mid s,a)
    =
    \prod_{j\in\mathcal{G}}
    p(s_j'\mid s_j,a).
    \label{eq:factorized_transition}
\end{equation}
An action affects the contamination state only within its contact footprint; cells outside that footprint are unchanged by the cleaning action. The dependence on $\theta_j$ allows regions with similar contamination levels to exhibit different stochastic responses to the same applied action.

\subsection{Planning Objective}
\label{sec:planning_objective}

The goal is to bring every region to the clean state,
\begin{equation}
    x_j=0,
    \qquad
    \forall j\in\mathcal{G}.
    \label{eq:cleaning_goal}
\end{equation}
Under the uncertain, spatially varying dynamics, our objective is to reach this state while balancing cleaning progress against redundant effort and travel. 
Because directly solving the full POMDP is generally intractable in this setting, we propose a closed-loop, belief space planner that adaptively and iteratively replans from observed surface outcomes.

\section{Approach}
\label{sec:approach}

Our approach couples a spatially-varying cleaning response model with a planner that constructs, evaluates, and adapts plans based on observations. Each planning iteration updates the belief state from the current surface observation, generates and scores candidate plans, executes the highest-scoring plan, and incorporates the resulting observation before replanning (\figref{fig:approach_overview}). The complete planning procedure is summarized in Appendix \ref{app:adaptive_cleaning}.

\begin{figure*}[t]
    \centering

    \newcommand{\plotimage}[2]{%
        \includegraphics[
            width=#1,
            trim=1bp 1bp 0bp 1bp,
            clip
        ]{#2}%
    }

    \begin{tikzpicture}[
        font=\small,
        >=Latex,
        panel/.style={
            align=center,
            inner sep=0pt
        },
        smallpanel/.style={
            align=center,
            inner sep=0pt
        },
        flow/.style={->, thick},
        feedback/.style={->, thick, rounded corners},
        arrowlabel/.style={font=\scriptsize, align=center, fill=white, inner sep=1pt},
        stage/.style={font=\small\bfseries, align=center},
        sublabel/.style={font=\scriptsize, align=center}
    ]

    \node[stage] at (0,3.2) {Surface Observation};
    \node[stage] at (3.15,3.2) {Belief Update};
    \node[stage] at (6.38,3.2) {Clustering};
    \node[stage] at (9.8,3.2) {Candidate Plans};
    \node[stage] at (13.6,3.2) {Post-Execution Surface};

    \node[panel] (sensor) at (0,0) {
        \plotimage{2.0cm}{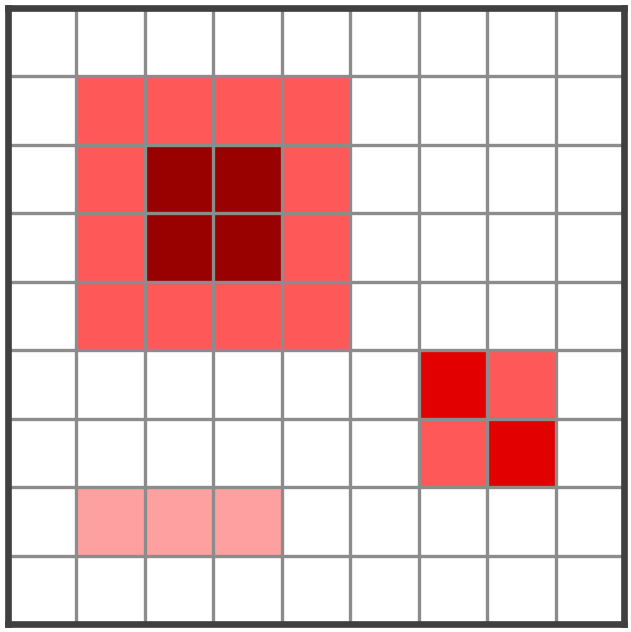}
    };

    \node[panel] (belief) at (3.2,0) {
        \plotimage{2.0cm}{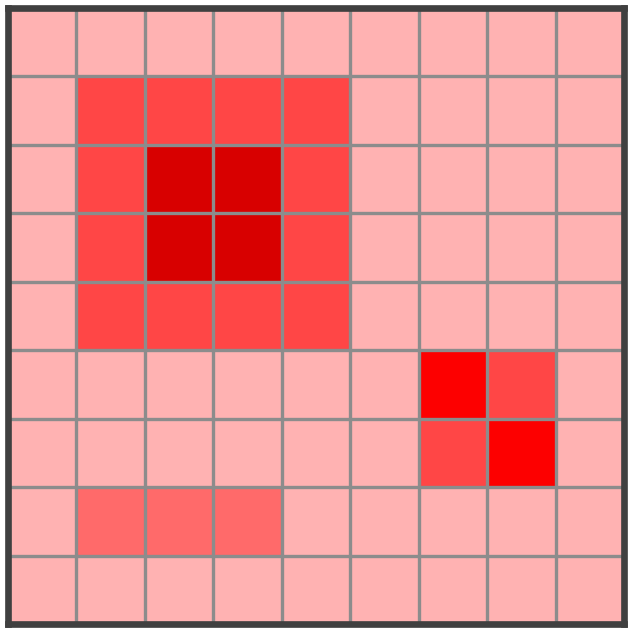}
    };

    \node[smallpanel] (cl1) at (6.43,2.05) {
        \plotimage{1.65cm}{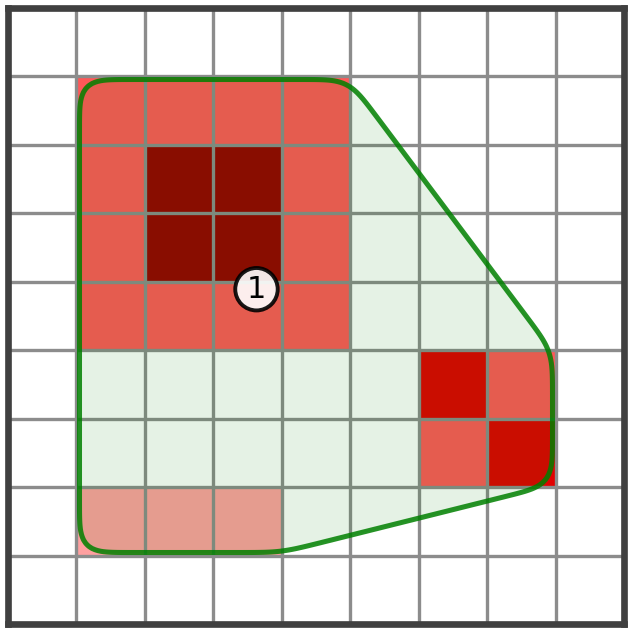}
    };
    \node[sublabel, anchor=north] at ($(cl1.south)+(-0.04,0.08)$) {$k=1$};

    \node[smallpanel] (cl2) at (6.43,0) {
        \plotimage{1.65cm}{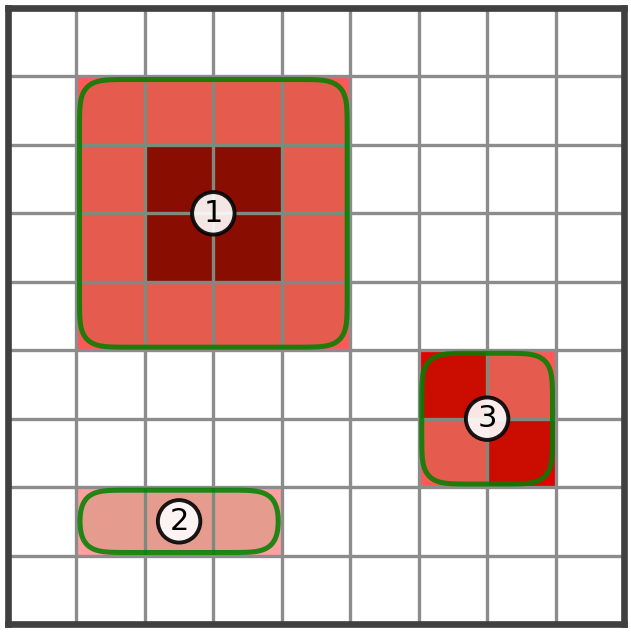}
    };
    \node[sublabel, anchor=north] at ($(cl2.south)+(-0.04,0.08)$) {$k=3$};

    \node[smallpanel] (cl3) at (6.43,-2.05) {
        \plotimage{1.65cm}{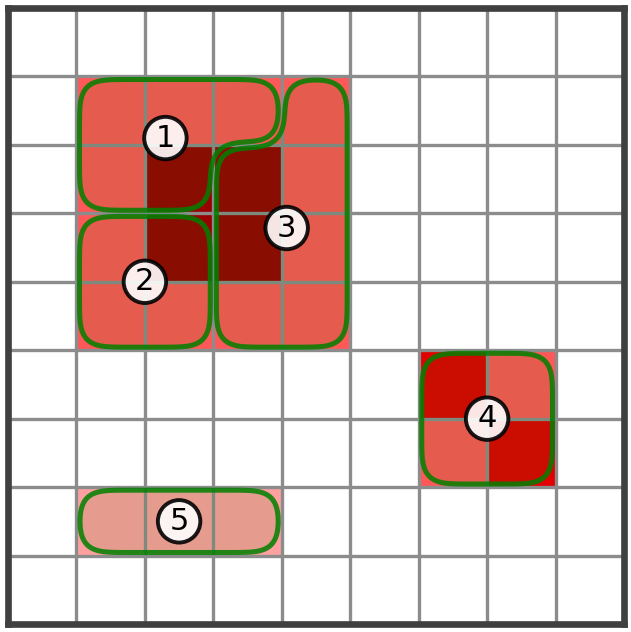}
    };
    \node[sublabel, anchor=north] at ($(cl3.south)+(-0.04,0.08)$) {$k=5$};

    \node[smallpanel] (cand1) at (9.8,2.05) {
        \plotimage{1.65cm}{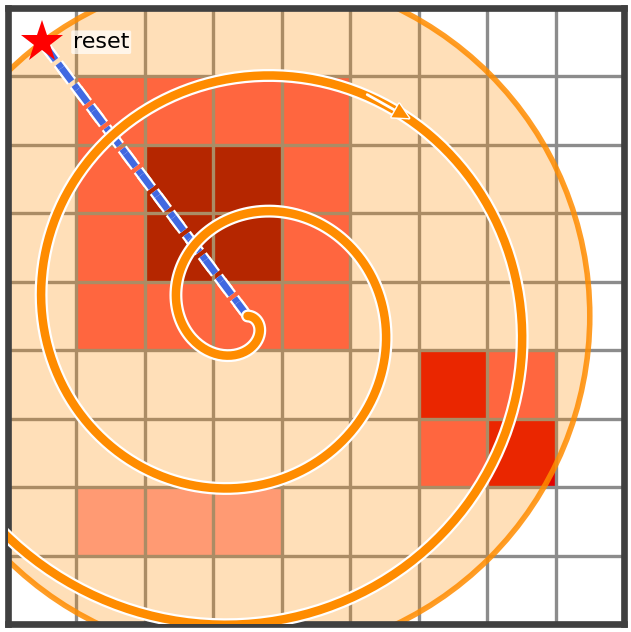}
    };
    \node[sublabel, anchor=north] at ($(cand1.south)+(0,0.08)$)
        {$J(P_1;b_t)=-15.62$};

    \node[smallpanel] (cand2) at (9.8,0.0) {
        \plotimage{1.65cm}{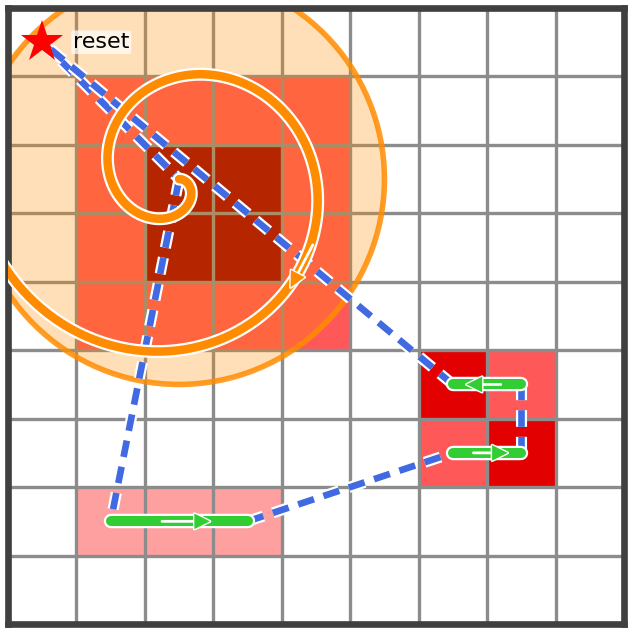}
    };
    \node[sublabel, anchor=north] at ($(cand2.south)+(0,0.08)$)
        {$J(P_2;b_t)=-12.14$};

    \node[smallpanel] (cand3) at (9.8,-2.05) {
        \plotimage{1.65cm}{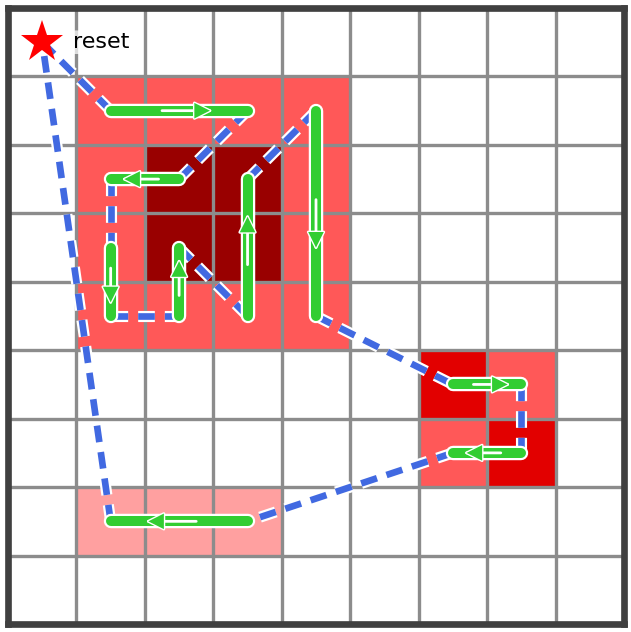}
    };
    \node[sublabel, anchor=north] at ($(cand3.south)+(0,0.08)$)
        {$J(P_3;b_t)=-12.73$};

    \node[panel] (exec) at (13.6,0) {
        \plotimage{2.0cm}{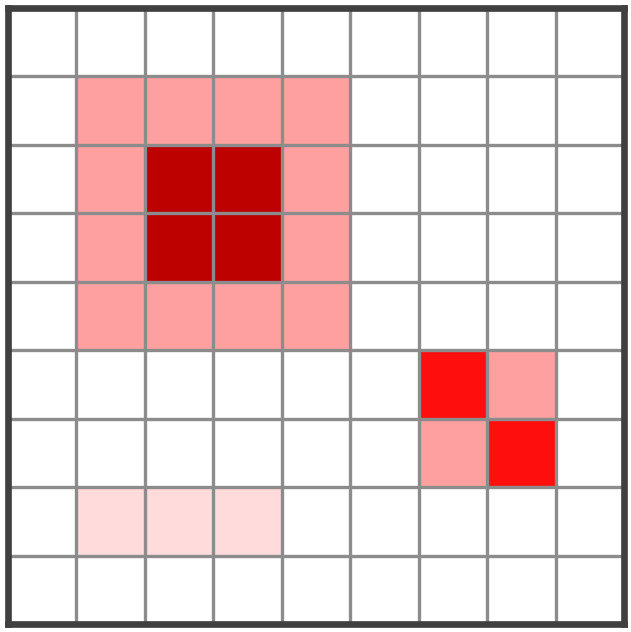}
    };

    \draw[flow]
        (sensor.east)
        --
        (belief.west);

    \draw[flow]
        (belief.east)
        --
        (cl1.west);

    \draw[flow]
        (belief.east)
        --
        (cl2.west);

    \draw[flow]
        (belief.east)
        --
        (cl3.west);

    \draw[flow]
        (cl1.east)
        --
        (cand1.west);

    \draw[flow]
        (cl2.east)
        --
        (cand2.west);

    \draw[flow]
        (cl3.east)
        --
        (cand3.west);

    \draw[flow]
        (cand2.east)
        --
        node[arrowlabel, above, yshift=2pt] {Selected for \\ Execution}
        (exec.west);

    \coordinate (feedback-right) at ($(exec.south)+(0,-2.50)$);
    \coordinate (feedback-left) at ($(sensor.south)+(0,-2.50)$);

    \draw[feedback]
        (exec.south)
        -- (feedback-right)
        -- node[arrowlabel, below]
        {Next iteration and new observation}
        (feedback-left)
        -- (sensor.south);

    \end{tikzpicture}

    \caption{Overview of the adaptive cleaning pipeline. At each iteration, the sensor input is converted into a scalar contamination observation to update the surface belief. Contaminated regions are grouped by spatial connectivity or alternative $k$-means clusterings (illustrated for $k=1,3,$ and $5$). For each cluster, the planner generates a candidate plan and evaluates $J(P;b_t)$ under the current belief. The highest-scoring plan is executed and then the updated surface is observed to begin the next cycle. Red shading indicates contamination (darker = higher $x_j$); shaded orange circles denote spiral actions, while green and blue line segments denote wipe and move actions, respectively.}
    \label{fig:approach_overview}
\end{figure*}

\subsection{Belief Modeling and Update}
\label{sec:belief_modeling}

True contamination dynamics depend on the contaminant, surface, and cleaning interaction and are generally unknown. Because directly estimating the latent response state $\theta_j$ is infeasible, we instead represent it with a scalar latent surrogate $h_j\in[0,1]$, which we refer to as \emph{stickiness} and interpret as an effective normalized force threshold. Rather than recovering the higher-dimensional $\theta_j$, $h_j$ captures a single force-dependent response characteristic for adapting cleaning actions.

We define the cell state and its belief mean as
\begin{equation}
    \xi_j=
    \begin{bmatrix}
        x_j\\
        h_j
    \end{bmatrix},
    \qquad
    \mu_j=
    \begin{bmatrix}
        \mu_{x,j}\\
        \mu_{h,j}
    \end{bmatrix}.
    \label{eq:internal_cell_state}
\end{equation}
For each cell, the planner maintains the joint Gaussian belief
\begin{equation}
    \xi_j
    \sim
    \mathcal{N}
    \left(
        \mu_j,
        P_j
    \right),
    \qquad
    P_j=
    \begin{bmatrix}
        \sigma_{x,j}^{2} & \sigma_{xh,j}\\
        \sigma_{xh,j} & \sigma_{h,j}^{2}
    \end{bmatrix}.
    \label{eq:joint_belief}
\end{equation}

\begin{figure*}[t]
    \centering

    \begin{minipage}[t]{0.24\textwidth}
        \centering
        \includegraphics[width=\linewidth]{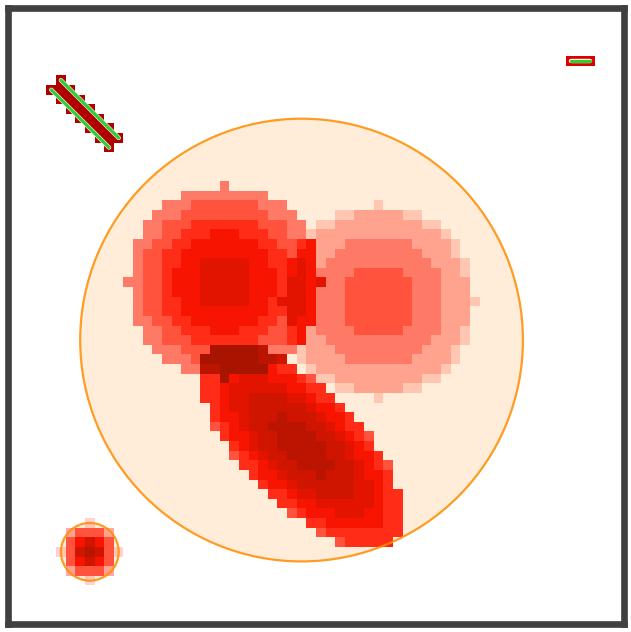}\\[-1mm]
        {\scriptsize (a) Iteration 0}
    \end{minipage}
    \hfill
    \begin{minipage}[t]{0.24\textwidth}
        \centering
        \includegraphics[width=\linewidth]{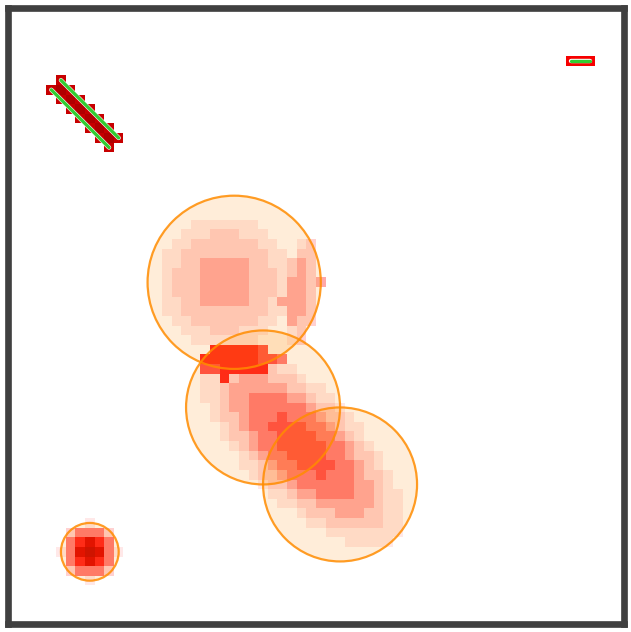}\\[-1mm]
        {\scriptsize (b) Iteration 1}
    \end{minipage}
    \hfill
    \begin{minipage}[t]{0.24\textwidth}
        \centering
        \includegraphics[width=\linewidth]{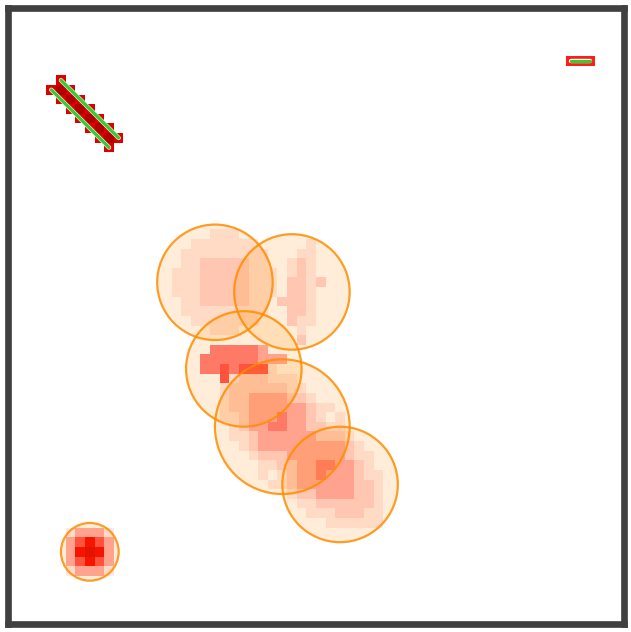}\\[-1mm]
        {\scriptsize (c) Iteration 2}
    \end{minipage}
    \hfill
    \begin{minipage}[t]{0.24\textwidth}
        \centering
        \includegraphics[width=\linewidth]{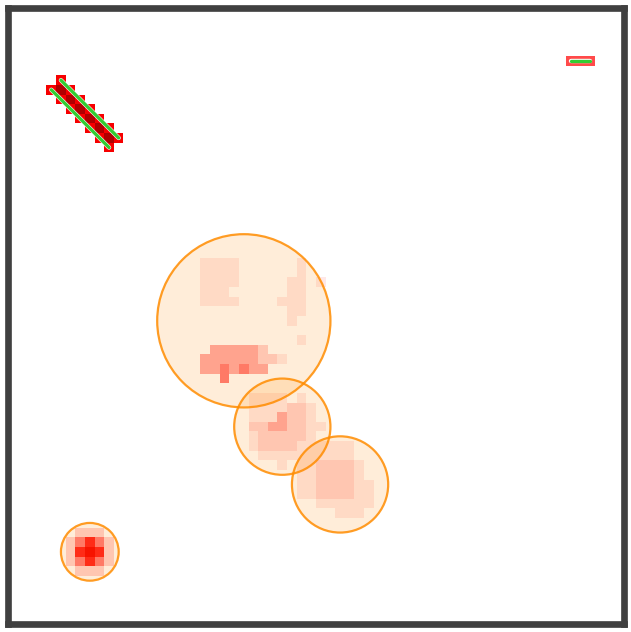}\\[-1mm]
        {\scriptsize (d) Iteration 3}
    \end{minipage}

    \vspace{1mm}

    \begin{minipage}[t]{0.24\textwidth}
        \centering
        \includegraphics[width=\linewidth]{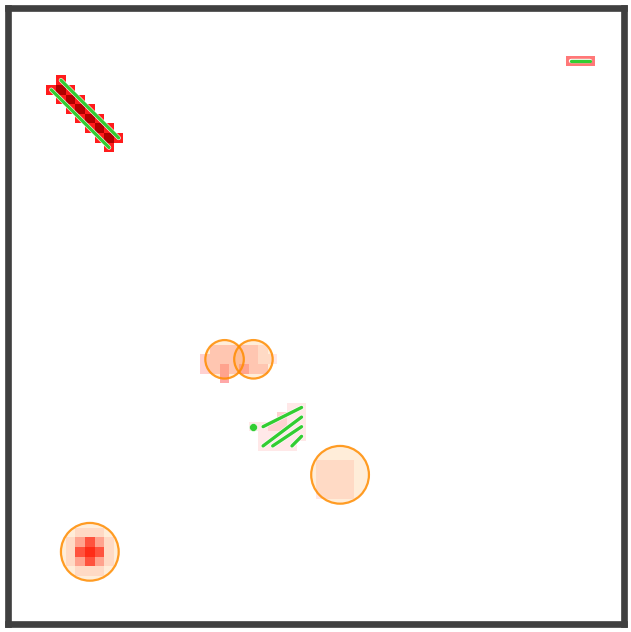}\\[-1mm]
        {\scriptsize (e) Iteration 4}
    \end{minipage}
    \hfill
    \begin{minipage}[t]{0.24\textwidth}
        \centering
        \includegraphics[width=\linewidth]{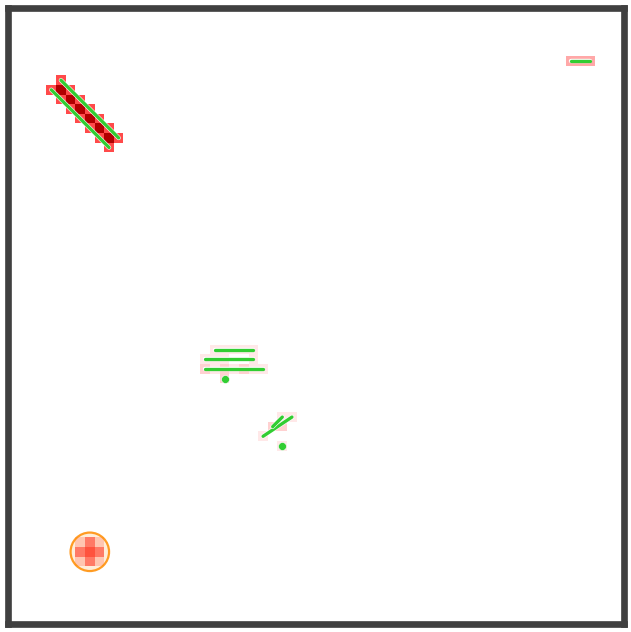}\\[-1mm]
        {\scriptsize (f) Iteration 5}
    \end{minipage}
    \hfill
    \begin{minipage}[t]{0.24\textwidth}
        \centering
        \includegraphics[width=\linewidth]{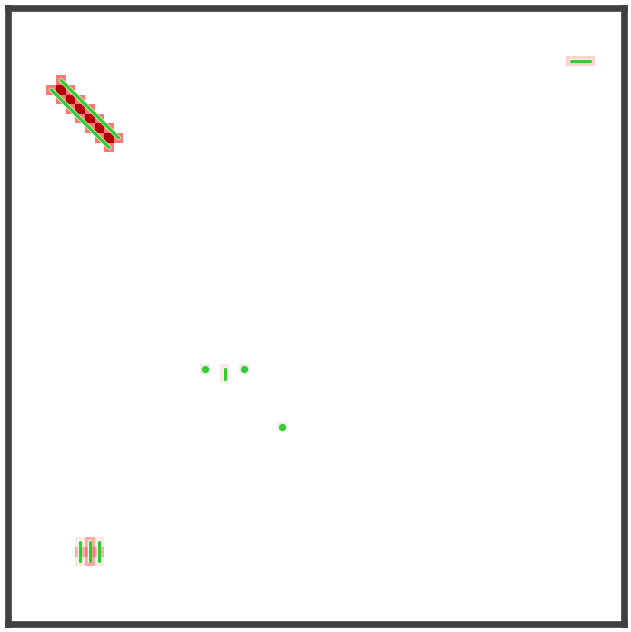}\\[-1mm]
        {\scriptsize (g) Iteration 6}
    \end{minipage}
    \hfill
    \begin{minipage}[t]{0.24\textwidth}
        \centering
        \includegraphics[width=\linewidth]{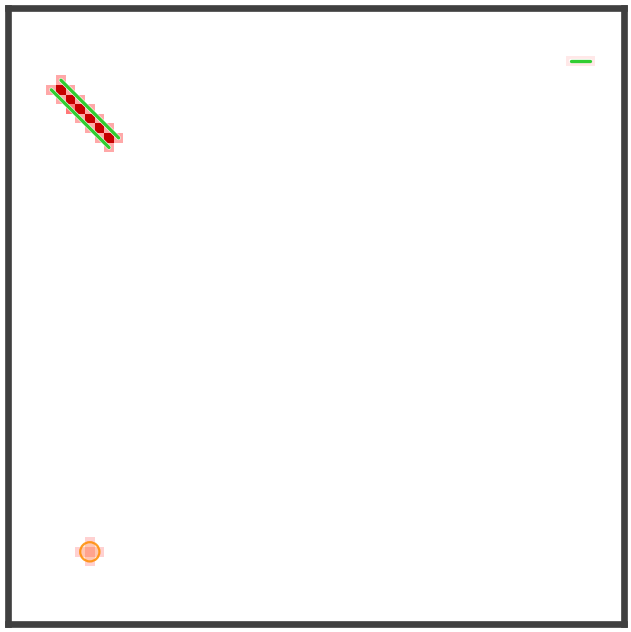}\\[-1mm]
        {\scriptsize (h) Iteration 7}
    \end{minipage}

    \caption{Evolution of the simulated surface and cleaning actions over planning iterations. Red shading indicates remaining contamination; orange circles denote spiral actions, while green outlines and line segments denote wipe actions and their trajectories. Panels (b)--(d) illustrate the planner's spatial decompositions: nearby contamination is grouped together in (b), split into clusters with separate actions in (c), and regrouped as the residual geometry changes in (d). In later iterations, sparse contamination leads the planner to favor local wipe actions over broad spiral actions to avoid unnecessary coverage of clean areas.}
    \label{fig:simulation_progress}
\end{figure*}

\subsubsection{Belief dynamics model}

The belief assumes an internal transition model capturing two properties motivated by everyday cleaning interactions. First, cleaning effectiveness depends nonlinearly on applied force: forces below a contamination-dependent response threshold may have little effect, while effectiveness increases once that threshold is exceeded. Second, each cleaning pass removes a fraction of the remaining contamination rather than a fixed absolute amount, reflecting how repeated wiping yields diminishing returns that leave residual contamination.

We model the fractional contamination reduction per pass using a smooth threshold response over normalized applied force $f_{norm}$:
\begin{equation}
    \alpha(f_{norm},h)
    =
    \alpha_{\max}w
    \ln\left(
        1+\exp\left(\frac{f_{norm}-h}{w}\right)
    \right),
    \label{eq:cleaning_response}
\end{equation}
with manually specified $\alpha_{\max}$ and $w$ corresponding to the removal scale and the smoothness of the force-response transition, respectively. With a per-pass residual fraction of
\begin{equation}
    q(f_{norm},h)=1-\alpha(f_{norm},h),
    \label{eq:survival_factor}
\end{equation}
the planner assumes the internal transition function of
\begin{equation}
    f_{\mathrm{int}}(\xi_j,a)
    =
    \begin{bmatrix}
        q(f_{norm},h_j)^k x_j\\
        h_j
    \end{bmatrix}.
    \label{eq:internal_transition_function}
\end{equation}
Hence the stochastic prediction is:
\begin{equation}
    \xi_j^-
    =
    f_{\mathrm{int}}(\xi_j,a)
    +
    \epsilon_j,
    \qquad
    \epsilon_j\sim\mathcal{N}(0,Q_j).
    \label{eq:ekf_transition_model}
\end{equation}
Thus, $h_j$ affects the predicted cleaning effectiveness but is unchanged by the transition itself. The process covariance $Q_j$ represents uncertainty.

The observation measures only the contamination coordinate. Let $\mathbf{e}_x=[1\;\;0]^\top$ denote the selector for that coordinate. The measurement model is:
\begin{equation}
    z_j
    =
    \mathbf{e}_x^\top \xi_j
    +v_j,
    \qquad
    v_j\sim\mathcal{N}(0,R).
    \label{eq:ekf_observation_model}
\end{equation}
Thus, $h_j$ is inferred from action outcomes rather than observed directly. The covariance parameters in $Q_j$ and $R$ are currently designed rather than learned.

\subsubsection{Belief update}

We use a per-cell extended Kalman filter (EKF) to maintain our belief because the transition in \eref{eq:internal_transition_function} is nonlinear in $h_j$. The EKF prediction is
\begin{equation}
    \mu_j^-
    =
    f_{\mathrm{int}}(\mu_j,a),
    \qquad
    P_j^-
    =
    F_jP_jF_j^\top+Q_j,
    \label{eq:ekf_prediction}
\end{equation}
where $F_j$ is the Jacobian of the internal transition model. The new appearance observation $z_j$ is then incorporated through the standard EKF measurement update:
\begin{equation}
    \mu_j^+
    =
    \mu_j^-+K_j\nu_j,
    \qquad
    P_j^+
    =
    (I-K_j\mathbf{e}_x^\top)P_j^-,
    \label{eq:ekf_update}
\end{equation}
where $\nu_j$ and $K_j$ are the innovation and Kalman gain, respectively. Appendix \ref{app:ekf} provides the complete prediction and measurement update equations.

Although observations directly measure only the contamination, action-dependent transitions create covariance between $x_j$ and $h_j$. Innovation therefore updates the latent response estimate: under-cleaned outcomes shift $h_j$ higher while over-effective cleaning lowers it. This outcome-driven update adapts subsequent cleaning decisions.

\subsection{Interleaved Manipulation Planning in Belief Space}
\label{sec:interleaved_planning}

At each iteration, the planner uses the current observation and belief to decide where and how to clean (\figref{fig:approach_overview}). To balance progress with efficiency, the planner constructs a set of candidate plans with varying spatial clustering and cleaning actions, then executes the highest-scoring plan under the current belief.

\subsubsection{Spatial Grouping and Candidate Generation}
\label{sec:spatial_grouping_candidate_generation}

To evaluate different spatial trade-offs, the planner generates alternative decompositions of observed contamination, $\mathcal{C}=\{C_1,\ldots,C_M\}$, grouping nearby regions when joint coverage is advantageous or separating them when targeted treatment is preferred.  For each cluster $C_m$, it generates a geometry-dependent sequence of cleaning actions: 
\begin{equation}
    \mathcal{A}_m
    =
    (a_{m,1},\ldots,a_{m,n_m}),
    \qquad
    a_{m,i}\in\mathcal{A},
    \label{eq:cluster_action_sequence}
\end{equation}
where $n_m$ is the number of actions. A cluster may require one action (e.g. a spiral for circular regions) or multiple actions (e.g. wipes joined by moves for elongated regions). 

The planner then searches over the force and pass parameters for each cleaning action, ordering and assembling them into a complete candidate plan $P$. See Appendix \ref{app:candidate_construction} for the detailed candidate-construction procedure.

\subsubsection{Belief-Aware Candidate Ranking}
\label{sec:belief_candidate_ranking}

Each candidate plan $P$ is scored by summing over $m$ clusters, given current belief $b$: 
\begin{equation}
    J(P;b)
    =
    \sum_{m=1}^{M}
    J(\mathcal{A}_m;b),
    \label{eq:plan_objective}
\end{equation}
where
\begin{align}
    J(\mathcal{A}_m;b)
    ={}&
    \lambda_{\mathrm{clean}}
    F_{\mathrm{clean}}(\mathcal{A}_m;b)
    -\lambda_{\mathrm{cost}}
    C(\mathcal{A}_m)
    \notag\\
    &-\lambda_{\mathrm{waste}}
    W(\mathcal{A}_m;b)
    -\lambda_{\mathrm{travel}}
    C_{\mathrm{travel}}(\mathcal{A}_m)
    \label{eq:cluster_objective}
\end{align}
penalizes action cost $C$, wasted effort $W$ and travel cost $C_{\mathrm{travel}}$ against predicted contamination reduction $F_{\mathrm{clean}}$, weighted by manually-set parameters $\lambda$ (Appendix \ref{app:objective}).

The planner evaluates all candidate plans, executes the highest-scoring plan, updates the belief from the observed outcomes, and repeats (\figref{fig:approach_overview}).

\section{Experiments and Results}
\label{sec:experiments}

We evaluate our planner's spatial decomposition and adaptive cleaning actions in a simplified simulator. 

\subsection{Simulation Setup}
\label{sec:simulation_setup}

The simulated surface is modeled by a $64\times64$  grid. Each cell stores a nonnegative integer state $X_j$, where $X_j=0$ is clean and higher values require greater effort to clean. Following \sref{sec:observation_model}, the initial observation fixes the normalization scale:
\begin{equation}
    X_{\max}
    =
    \max\left(1,\max_{j\in\mathcal{G}}X_j^{(0)}\right),
    \label{eq:simulation_obs_scale}
\end{equation}
and subsequent observations directly provide the scalar contamination $z_{j}$:
\begin{equation}
    z_j
    =
    \min\left(\frac{X_j}{X_{\max}},1\right).
    \label{eq:simulation_observation}
\end{equation}

The simulator transition function is deterministic, with cleaning actions reducing cell states based on an action and parameter dependent scalar decrement. The simulator does not model the latent response variable $h_j$; instead, $h_j$ is used purely by the planner to account for differences between predicted and observed cleaning outcomes. The experiment terminates when $X_j=0$ for all cells.

\subsection{Preliminary Results}
\label{sec:preliminary_results}

We consider several initial contamination geometries, including compact and overlapping regions, elongated structures, and spatially separated contamination, to exercise different spatial decompositions and cleaning primitives. \figref{fig:simulation_progress} shows eight consecutive planning iterations for one representative contamination map. As cleaning progresses, the planner adapts its spatial decomposition and action selection, penalizing broad actions once the remaining contamination becomes sparse to avoid wasted effort over clean areas.

This sequence provides qualitative evidence of the planner's online adaptability. Future work will extend this evaluation to real-world experiments with visual observations and robot-executed cleaning actions.

\clearpage
\bibliographystyle{IEEEtran}
\bibliography{reference}

\appendices

\section{Action Space and Cost Functions}
\label{app:action_definitions}

The concrete parameterization of the action space is summarized in \tref{tab:action_space}. Spatial parameters are expressed in grid coordinates, while cleaning parameters are selected from finite sets. Here, $p_s$ and $p_e$ denote the start and end positions, $p_c$ denotes the spiral center, $r$ denotes the spiral radius, $f$ denotes the force level, and $k$ denotes the number of passes. One pass constitutes a single endpoint-to-endpoint traversal for a scratch action and a complete loop for a spiral action. The force corresponding to each discrete force level is manually specified.

\begin{table}[h]
    \centering
    \caption{Action Space}
    \label{tab:action_space}
    \begin{tabular}{lll}
        \hline
        Action & Spatial parameters & Cleaning parameters \\
        \hline
        Wipe &
        $(p_s,p_e)\in\mathcal{G}^{2}$ &
        $f\in\{1,2,3,4,5\}$ \\
        Scratch &
        $(p_s,p_e)\in\mathcal{G}^{2}$ &
        $f\in\{1,2,3,4,5\},\ k\in\{1,2,3\}$ \\
        Spiral &
        $p_c\in\mathcal{G},\ r\in\mathbb{Z}_{>0}$ &
        $f\in\{1,2,3,4,5\},\ k\in\{1,2,3\}$ \\
        Move &
        $(p_s,p_e)\in\mathcal{G}^{2}$ &
        --- \\
        \hline
    \end{tabular}
\end{table}
Defining the force-dependent factor
\begin{equation}
    c_f(f)
    =
    \exp\left(\frac{f-1}{5}\right),
    \label{eq:force_cost_factor}
\end{equation}
we let
\begin{equation}
    d(p_s,p_e)
    =
    \lVert p_e-p_s\rVert_2,
    \label{eq:grid_action_distance}
\end{equation}
where $\lVert \cdot \rVert_2$ is the Euclidean distance in grid coordinates. Using these quantities, the action-specific cost functions used in candidate-plan evaluation are summarized in \tref{tab:action_cost}.

\begin{table}[h]
    \centering
    \caption{Cost Functions}
    \label{tab:action_cost}
    \begin{tabular}{ll}
        \hline
        Action & Cost $c(a)$ \\
        \hline
        Wipe &
        $c_f(f) w_{\mathrm{len}}\left(d(p_s,p_e)+1\right)$ \\
        Scratch &
        $c_f(f) k w_{\mathrm{len}}\left(d(p_s,p_e)+1\right)$ \\
        Spiral &
        $c_f(f) k\left(2\pi r w_{\mathrm{len}}
        +(\pi r^2)^2w_{\mathrm{area}}\right)$ \\
        Move &
        $c_{\mathrm{move}}d(p_s,p_e)$ \\
        \hline
    \end{tabular}
\end{table}

The manually specified weights $w_{\mathrm{len}}$, $w_{\mathrm{area}}$, and
$c_{\mathrm{move}}$ balance path length, covered area, and travel distance,
respectively. For linear cleaning actions, the $+1$ term accounts for the
starting cell.

\section{Adaptive Planning Procedure}
\label{app:adaptive_cleaning}

The complete closed-loop planning procedure is summarized in \aref{alg:adaptive_cleaning}. At each iteration, the planner updates its belief from the latest surface observation, constructs candidate plans from alternative spatial groupings, scores these candidates under the current belief, and executes the highest-scoring plan before observing the resulting surface and replanning. Candidate construction is detailed separately in Appendix \ref{app:candidate_construction}.

\begin{algorithm}[t]
\caption{Belief-Aware Adaptive Cleaning}
\label{alg:adaptive_cleaning}
\begin{algorithmic}[1]
\STATE Initialize belief $b_0$ and previous plan $P_{\mathrm{prev}}\leftarrow\varnothing$
\WHILE{surface is not clean}
    \STATE $z \leftarrow \textsc{ObserveSurface()}$  
    \IF{$P_{\mathrm{prev}}\neq\varnothing$}
        \FOR{$a\in P_{\mathrm{prev}}$}
            \STATE $b \leftarrow \textsc{EKFBeliefPredict}(b,a)$
        \ENDFOR
    \ENDIF
    \STATE $b \leftarrow \textsc{EKFBeliefUpdate}(b,z)$
    \STATE $\mathcal{C}_{\mathrm{cc}}\leftarrow \textsc{Generate8ConnectedCluster()}$ 
    \STATE $P_{\mathrm{cc}}\leftarrow\textsc{ConstructCandidate}(\mathcal{C}_{\mathrm{cc}},b)$
    \STATE $J_{\mathrm{cc}}\leftarrow J(P_{\mathrm{cc}};b)$

    \STATE $P_{\mathrm{km}}^\star\leftarrow\varnothing$, $J_{\mathrm{km}}^\star\leftarrow-\infty$
    \FOR{$k=1,2,\ldots$}
        \STATE $\mathcal{C}_k\leftarrow\textsc{WeightedKMeans}(z,b,k)$
        \STATE $P_k\leftarrow\textsc{ConstructCandidate}(\mathcal{C}_k,b)$
        \STATE $J_k\leftarrow J(P_k;b)$
        \IF{$J_k>J_{\mathrm{km}}^\star$}
            \STATE $(P_{\mathrm{km}}^\star,J_{\mathrm{km}}^\star)\leftarrow(P_k,J_k)$
        \ENDIF
        \IF{$k>2$ \AND $J_k\leq J_{k-1}\leq J_{k-2}$}
            \STATE \textbf{break}
        \ENDIF
    \ENDFOR

    \STATE $P^\star\leftarrow\arg\max_{P\in\{P_{\mathrm{cc}},P_{\mathrm{km}}^\star\}}J(P;b)$
    \STATE $\textsc{Execute}(P^\star)$
    \STATE $P_{\mathrm{prev}}\leftarrow P^\star$
\ENDWHILE
\end{algorithmic}
\end{algorithm}

\section{Extended Kalman Filter}
\label{app:ekf}

The planner maintains the per-cell internal state
\begin{equation}
    \xi_j=
    \begin{bmatrix}
        x_j\\
        h_j
    \end{bmatrix},
\end{equation}
with internal transition
\begin{equation}
    f_{\mathrm{int}}(\xi_j,a)
    =
    \begin{bmatrix}
        q(f_{\mathrm{norm}},h_j)^k x_j\\
        h_j
    \end{bmatrix},
\end{equation}
where
\begin{equation}
    q(f_{\mathrm{norm}},h)=1-\alpha(f_{\mathrm{norm}},h).
\end{equation}
Following standard EKF derivation, predicted mean is obtained by evaluating the internal transition function at the current belief mean,
\begin{equation}
    \mu_j^-
    =
    f_{\mathrm{int}}(\mu_j,a).
    \label{eq:app_ekf_mean_prediction}
\end{equation}
In particular,
\begin{equation}
    \mu_{x,j}^-
    =
    q(f_{norm},\mu_{h,j})^k\mu_{x,j},
    \qquad
    \mu_{h,j}^-=\mu_{h,j}.
    \label{eq:app_ekf_mean_prediction_explicit}
\end{equation}
The transition Jacobian is
\begin{equation}
    F_j
    =
    \left.
    \frac{\partial f_{\mathrm{int}}(\xi_j,a)}
         {\partial \xi_j}
    \right|_{\xi_j=\mu_j}
    =
    \begin{bmatrix}
        q_j^k & F_{xh,j}\\
        0 & 1
    \end{bmatrix},
    \label{eq:app_ekf_jacobian}
\end{equation}
where $q_j=q(f_{norm},\mu_{h,j})$ and
\begin{equation}
    F_{xh,j}
    =
    -kq_j^{k-1}\mu_{x,j}
    \left.
    \frac{\partial\alpha(f_{norm},h)}{\partial h}
    \right|_{h=\mu_{h,j}}.
    \label{eq:app_ekf_cross_jacobian}
\end{equation}
The covariance prediction follows the standard EKF update
\begin{equation}
    P_j^-
    =
    F_jP_jF_j^\top+Q_j.
    \label{eq:app_ekf_cov_prediction}
\end{equation} where $Q_j$ is the process-noise covariance.

The belief-update stage incorporates the new appearance observation $z_j$. The observation measures the contamination coordinate, with
\begin{equation}
    \mathbf{e}_x=
    \begin{bmatrix}
        1\\
        0
    \end{bmatrix}.
\end{equation} The innovation and innovation covariance are
\begin{equation}
    \nu_j=z_j-\mathbf{e}_x^\top\mu_j^-,
    \qquad
    S_j=\mathbf{e}_x^\top P_j^-\mathbf{e}_x+R.
    \label{eq:app_ekf_innovation}
\end{equation} where $R$ is the observation-noise variance.
The Kalman gain is
\begin{equation}
    K_j
    =
    P_j^-\mathbf{e}_x S_j^{-1},
    \label{eq:app_ekf_gain}
\end{equation}
and the posterior belief is
\begin{align}
    \mu_j^+
    &=
    \mu_j^-+K_j\nu_j,
    \label{eq:app_ekf_mean_update}\\
    P_j^+
    &=
    (I-K_j\mathbf{e}_x^\top)P_j^-.
    \label{eq:app_ekf_cov_update}
\end{align}

\section{Spatial Grouping and Candidate Generation}
\label{app:candidate_construction}

This section provides the clustering and candidate-construction details for the procedure summarized in \sref{sec:spatial_grouping_candidate_generation}. The current implementation generates alternative clusterings $\mathcal{C}$ using both 8-connected components and belief-weighted $k$-means clustering of contaminated cells. The 8-connected-component clustering separates spatially disconnected contamination, while the $k$-means candidates provide alternative partitions that may merge nearby regions or divide larger regions into multiple clusters.

For the $k$-means candidates, the contaminated-cell coordinates are clustered using the current contamination belief $\mu_{x,j}$ as weights. Multiple values of $k$ are considered, producing alternative clusterings of the same observed contamination. Together with the 8-connected-component clustering, these alternatives define the candidate clusterings considered by the planner.

Given a clustering $\mathcal{C}=\{C_1,\ldots,C_M\}$, \aref{alg:candidate_construction} summarizes how the corresponding cluster action sequences $\mathcal{A}_m$ are generated, parameterized, and assembled into a candidate plan $P$.

\begin{algorithm}[t]
\caption{\textsc{ConstructCandidate}$(\mathcal{C},b)$}
\label{alg:candidate_construction}
\begin{algorithmic}[1]
\FOR{each cluster $C_m\in\mathcal{C}$}
    \STATE $\mathcal{A}_m \leftarrow \varnothing$
    \STATE $p_m \leftarrow \textsc{ComputePCA}(C_m)$
    \IF{$\textsc{IsCircular}(p_m)$}
        \STATE $a \leftarrow \textsc{GenerateSpiralSkeleton}(C_m,p_m)$
    \ELSE
        \STATE $a \leftarrow \textsc{GenerateWipeSkeleton}(C_m,p_m)$
    \ENDIF
    \STATE $\mathcal{A}_m \leftarrow \mathcal{A}_m \cup \{a\}$
\ENDFOR
\STATE $A \leftarrow \textsc{GreedyNearestNeighbor}(\{\mathcal{A}_m\}_{m=1}^{M})$
\FOR{each $a_i\in A$}
    \STATE $(a_i^{\mathrm{force}},a_i^{\mathrm{pass}})
    \leftarrow \textsc{OptimizeParameters}(a_i,b)$
\ENDFOR
\STATE $P \leftarrow \textsc{AssemblePlanWithTraversals}(A)$
\RETURN $P$
\end{algorithmic}
\end{algorithm}

For each cluster $C_m$, principal component analysis (PCA) is applied to the coordinates of its contaminated cells to characterize the cluster geometry. The relative magnitudes of the principal axes are used to distinguish approximately circular from elongated clusters and to construct the corresponding action sequence $\mathcal{A}_m$ introduced in \sref{sec:spatial_grouping_candidate_generation}. Approximately circular clusters produce spiral candidates, whereas elongated clusters are covered by PCA-aligned wipe actions; multiple wipe actions may be required to cover a single cluster. Small or degenerate clusters are handled using wipe primitives. The resulting cluster action sequences are then ordered and parameterized as summarized in \aref{alg:candidate_construction}.

\section{Planning Objective}
\label{app:objective}

This section provides the detailed objective terms used in the belief-aware candidate ranking described in \sref{sec:belief_candidate_ranking}. Recall that a candidate plan $P$ is composed of cluster action sequences $\mathcal{A}_m$, each of which is evaluated using the objective in \eref{eq:cluster_objective}.

Let $\mathcal{A}_m^{\mathrm{clean}}\subseteq\mathcal{A}_m$ denote the contact cleaning actions in the sequence, excluding move actions. The predicted cleaning effect is evaluated jointly over all cleaning actions associated with the cluster,
\begin{equation}
    F_{\mathrm{clean}}(\mathcal{A}_m;b)
    =
    \sum_j
    \mu_{x,j}
    \left[
        1-
        \prod_{\substack{
            a\in\mathcal{A}_m^{\mathrm{clean}}:\\
            j\in\operatorname{cov}(a)
        }}
        q(f_a,\mu_{h,j})^{k_a}
    \right],
    \label{eq:current_cleaning_term}
\end{equation}
where $q(f_a,\mu_{h,j})$ is the predicted residual contamination fraction for action $a$. Joint evaluation composes the effects of multiple actions covering the same cell and avoids independently crediting overlapping action footprints.

The cleaning cost is additive over the contact cleaning actions,
\begin{equation}
    C(\mathcal{A}_m)
    =
    \sum_{a\in\mathcal{A}_m^{\mathrm{clean}}}
    c(a;\gamma_a),
    \label{eq:cluster_action_cost}
\end{equation}
where $\gamma_a$ denotes the parameters of action $a$, including its geometry, applied force, and pass count. The current implementation assigns greater cost to longer motions, larger applied force, and additional passes. The precise functional form is a current design choice rather than a calibrated physical effort model.

The waste term accumulates action-dependent effort according to the probability that covered cells are already below a contamination threshold,
\begin{equation}
    W(\mathcal{A}_m;b)
    =
    \sum_{a\in\mathcal{A}_m^{\mathrm{clean}}}
    c(a;\gamma_a)
    \frac{1}{|\operatorname{cov}(a)|}
    \sum_{j\in\operatorname{cov}(a)}
    \Phi\left(
        \frac{\tau-\mu_{x,j}}
             {\sigma_{x,j}}
    \right),
    \label{eq:current_waste_term}
\end{equation}
where $\operatorname{cov}(a)$ denotes the grid cells covered by action $a$, $\Phi$ is the standard normal cumulative distribution function, and $\tau$ is the current clean-region threshold.

The travel term $C_{\mathrm{travel}}(\mathcal{A}_m)$ accounts for the move actions required to enter the cluster, connect cleaning actions within the cluster, and transition toward the subsequent cluster. Together, these terms define the cluster score $J(\mathcal{A}_m;b)$ in \eref{eq:cluster_objective}. 

During candidate construction, the planner searches over the available force and pass parameters and selects the parameterization maximizing this score for each cluster, as summarized in \aref{alg:candidate_construction}. The resulting candidate plans are compared using the total score in \eref{eq:plan_objective}, and the highest-scoring candidate is selected for execution. The specific functional forms and relative weights used in the current implementation remain subject to calibration and refinement.

\end{document}